\documentclass[10pt,twocolumn,letterpaper]{article}
\usepackage[accsupp]{axessibility}  
\usepackage{iccv}
\usepackage{times}
\usepackage{epsfig}
\usepackage{graphicx}
\usepackage{amsmath}
\usepackage{amssymb}
\usepackage{booktabs}
\usepackage{makecell}
\DeclareMathOperator{\Tr}{Tr}
\usepackage[table]{xcolor}
\usepackage{multirow}

\usepackage[pagebackref=true,breaklinks=true,letterpaper=true,colorlinks,bookmarks=false]{hyperref}

\iccvfinalcopy 

\ificcvfinal\fi

\begin{document}

\title{EventHPE: Event-based 3D Human Pose and Shape Estimation}

\author{
Shihao Zou$^{\ast}$ \qquad Chuan Guo$^{\ast}$  \qquad Xinxin Zuo$^{\ast,\S}$  \qquad Sen Wang$^{\ast,\S}$  \qquad Pengyu Wang$^{\ddagger}$  \\ Xiaoqin Hu$^{\dagger}$  \qquad Shoushun Chen$^{\dagger, \natural}$  \qquad Minglun Gong$^{\S}$  \qquad Li Cheng$^{\ast}$\\
$^{\ast}$University of Alberta \qquad $^{\ddagger}$Shandong University \qquad $^{\dagger}$Celepixel Technology \\$^{\S}$University of Guelph \qquad $^{\natural}$ Nanyang Technological University\\
\tt\small \{szou2, cguo2, xzuo, sen9, lcheng5\}@ualberta.ca, pengyuwang@mail.sdu.edu.cn, \\ \tt\small \{xiaoqin.hu, shoushun.chen\}@ovt.com, minglun@uoguelph.ca
}

\maketitle
\ificcvfinal\thispagestyle{empty}\fi

\begin{abstract}
   Event camera is an emerging imaging sensor for capturing dynamics of moving objects as events, which motivates our work in estimating 3D human pose and shape from the event signals. Events, on the other hand, have their unique challenges: rather than capturing static body postures, the event signals are best at capturing local motions. This leads us to propose a two-stage deep learning approach, called EventHPE. The first-stage, FlowNet, is trained by unsupervised learning to infer optical flow from events. Both events and optical flow are closely related to human body dynamics, which are fed as input to the ShapeNet in the second stage, to estimate 3D human shapes. To mitigate the discrepancy between image-based flow (optical flow) and shape-based flow (vertices movement of human body shape), a novel flow coherence loss is introduced by exploiting the fact that both flows are originated from the identical human motion. An in-house event-based 3D human dataset is curated that comes with 3D pose and shape annotations, which is by far the largest one to our knowledge. Empirical evaluations on DHP19 dataset and our in-house dataset demonstrate the effectiveness of our approach. 
\end{abstract}

\section{Introduction}

\begin{figure}[htb]
    \centering
    \includegraphics[width=0.9\columnwidth]{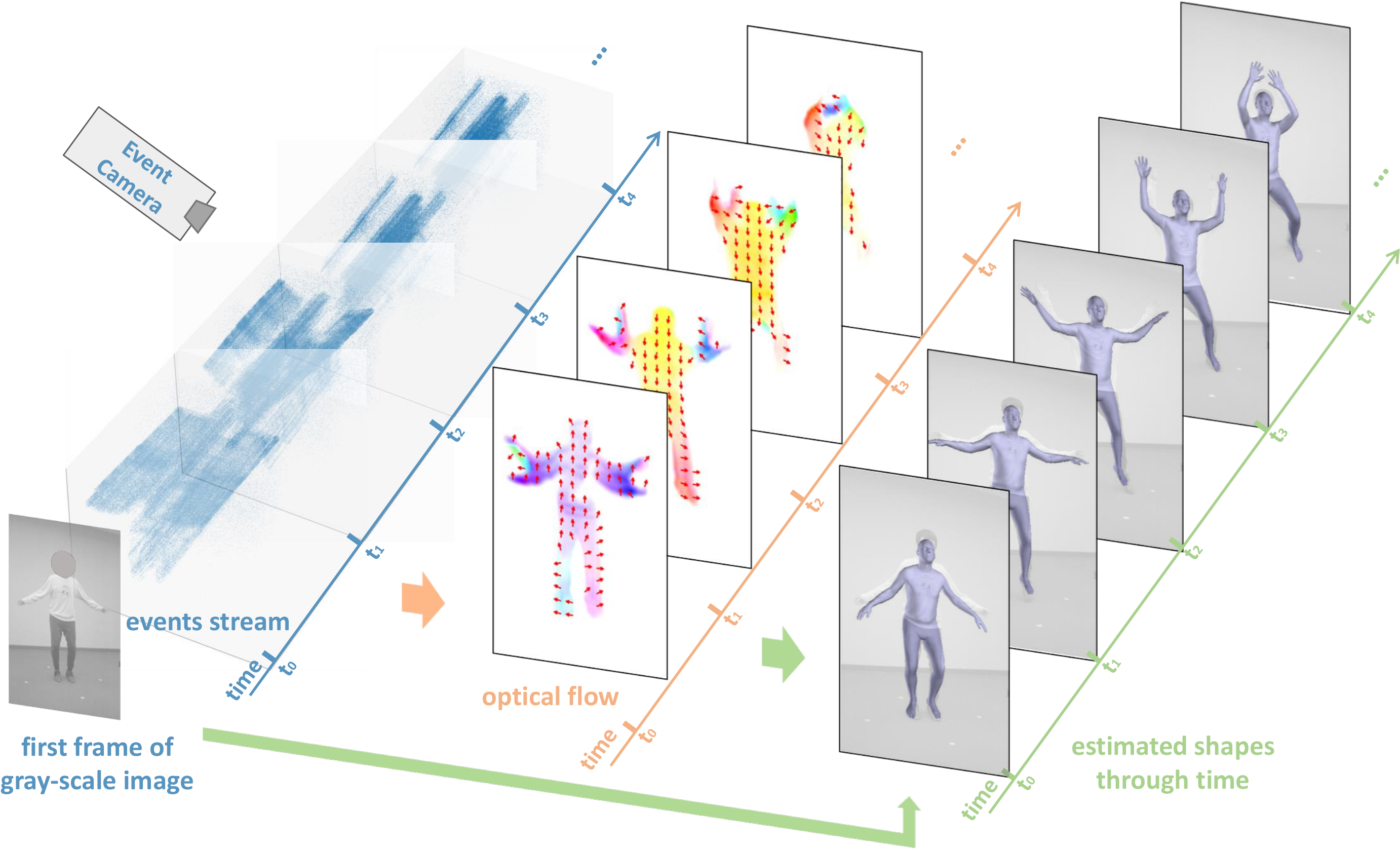}
    \caption{An illustration of our approach, which takes as input a stream of events and \emph{only} the first frame of gray-scale image from an event camera. Our goal is to estimate 3D human poses and shapes through time from the events stream as the sole data source given the beginning body posture in the first frame of gray-scale image, where optical flows are inferred from events as an intermediate step.}
    \label{fig:overview}
\end{figure}

Visual human pose and shape estimation have played a critical role in computer vision, with numerous research activities over the years~\cite{CVIU01:MoeGra,CVIU16:SarEtAl}.
Existing research efforts are predominantly based on images from the conventional RGB or RGB-D cameras~\cite{kanazawa2018end,kocabas2019vibe,humanMotionKanazawa19,xu2019denserac,zhu2021detailed}. 
Meanwhile, the recent development in event cameras~\cite{gallego2019event} offers new opportunities. Its working mechanism is a paradigm shift from theses conventional frame-based cameras. 
Inspired by biological vision process, event cameras~\cite{gallego2019event} asynchronously measure per-pixel brightness changes, which enables them to best detect and capture object local motions. 
This new imaging paradigm has already stimulated a range of computer vision research activities, such as camera pose estimation~\cite{gallego2017event}, gesture recognition~\cite{amir2017low} and 3D reconstruction~\cite{rebecq2018emvs}, 
as well as commercial interests spanning a range of use scenarios including robotics, augmented and virtual reality, and auto-driving applications~\cite{gallego2019event}.
However, its potential in estimating 3D human shape is rarely explored. 


DHP19~\cite{calabrese2019dhp19} is an early work that estimates only 2-D poses by treating a packet of events as a static image. 
The recent effort of EventCap~\cite{xu2020eventcap} is the first in estimating 3D human shapes from an event camera. 
However, in addition to the input events, EventCap relies on an additional stream of gray-scale image sequence as input, to establish the initial shape estimation at each time step. 
This motivates us to consider the problem of inferring 3D human shape from events as the major source of input: events are the sole source of input data to estimate 3D human shapes over time, given that the beginning shape is known or extracted from the first frame of gray-scale image.
Fig.~\ref{fig:overview} provides an overview of our two-stage approach, called EventHPE. 


Considering the fact that the two modalities, events and optical flow, are both closely related to human motions, and optical flow can provide explicit geometric information to describe human body movements, we place the inference of optical flow from events (i.e. FlowNet) in the first stage of our framework, which is trained without supervision. The inclusion of optical flow makes it feasible to estimate human poses and shapes mainly from events, which means we do not require a stream of gray-scale images as input in addition to the events.
The second stage, denoted by ShapeNet, is to estimate shape variations over time, given the events and the inferred optical flows as input. 
A novel flow coherence loss is proposed to enforce consistency of image-based flow (optical flow) and shape-based flow (vertices movement of human body shape), as both modalities are originated from the same human motion. 
Our main contributions are summarized below. (1) We present an approach to a new and challenging problem, estimating 3D human parametric shape mainly from events. We propose to leverage optical flow inferred from events to relieve the reliance on the gray-scale image sequence as additional input. A novel coherence loss is also introduced to ensure consistency between image-based flow (optical flow) and shape-based flow (vertices movement of human body shape). Empirical evaluations demonstrate the superior performance of our approach against several state of the arts. (2) A home-grown dataset is introduced, referred as Multi-Modality Human Pose and Shape Dataset (or MMHPSD)~\footnote{Our code and dataset are at \href{https://github.com/JimmyZou/EventHPE}{https://github.com/JimmyZou/EventHPE}.}. It includes 240k frames, with each frame containing 12 images from multiple imaging modalities including event camera. To our knowledge, MMHPSD is the largest event-based 3D human pose and shape dataset, and is the first publicly available dataset of such type, since the dataset of EventCap~\cite{xu2020eventcap} is not publicly available. The multi-modality property of MMHPSD also renders its great potential in facilitating existing and new research directions.

\begin{figure*}
    \centering
    \includegraphics[width=0.95\textwidth]{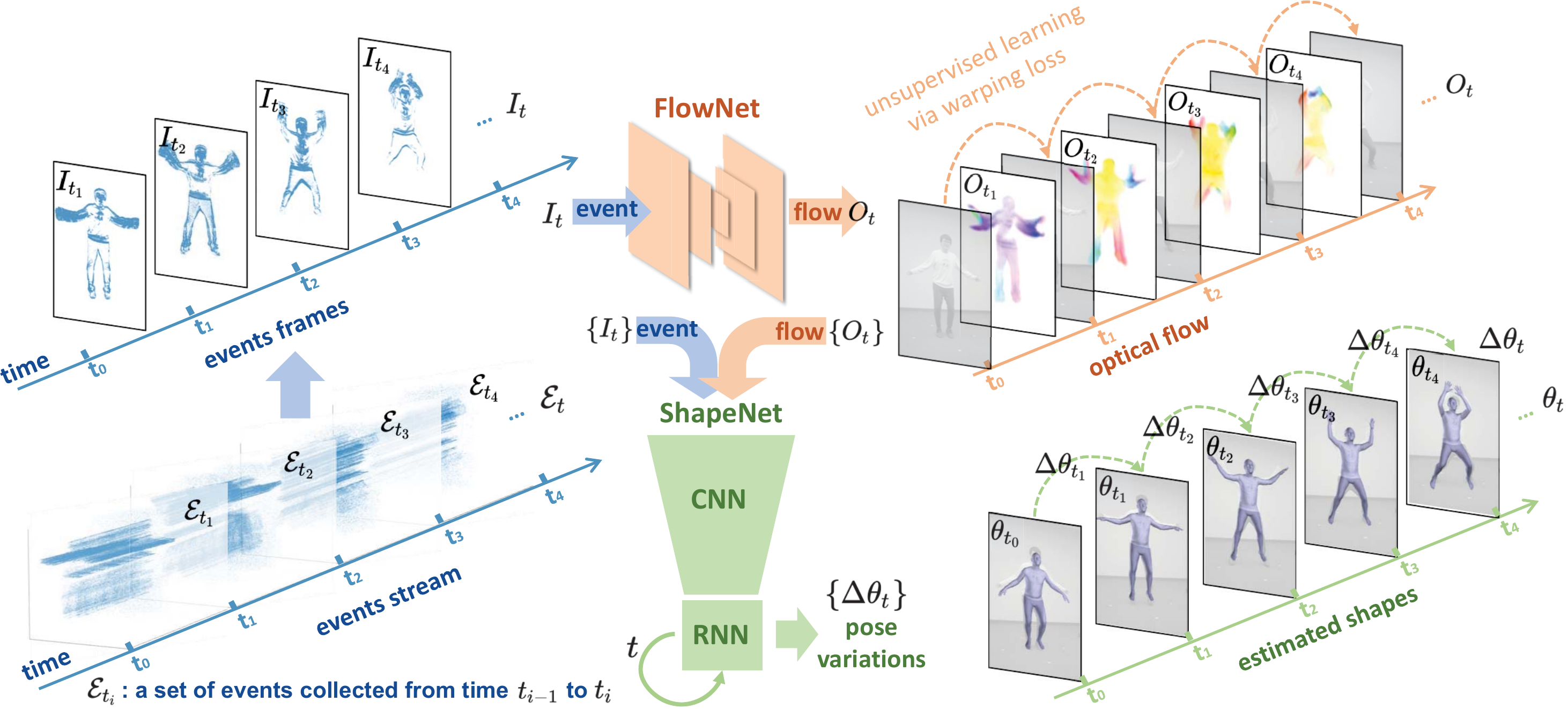}
    \caption{An overview of our EventHPE framework that consists of two stages. Events within time interval $(t_{i-1}, t_{i})$ are accumulated as an events packet $\mathcal{E}_{t_i}$, and then aggregated to an event frame $I_{t_i}$. (i) In stage one, our FlowNet infers optical flow from events, where event frames $I_{t_i}$ are fed into a CNN model to predict the optical flow $O_{t_i}$. (ii) At stage two, our ShapeNet takes in a sequence of event frames and its corresponding optical flows. A CNN module is used to extract the vectorized feature representation, which are passed to an RNN module to infer pose variations $\Delta \boldsymbol{\theta}_{t_i}$ during time interval $(t_{i-1}, t_{i})$. After articulating the beginning pose $\boldsymbol{\theta}_{t_0}$ and shape $\beta$, the body shape at each time point $t_i$ are subsequently estimated.}
    \label{fig:pipeline}
\end{figure*}

\section{Related Work}
\textbf{Human pose and shape estimation.} 
Human pose estimation from RGB or depth images has been extensively investigated in the past few years. The approaches prior to the deep learning era are primary dictionary learning-based~\cite{wang2014robust,zhou2016sparseness,chen20173d}. This is followed by the wide-spread use of deep learning techniques with noticeable performance increments, include e.g. direct regression of 3D pose~\cite{park20163d,li2015maximum}, or lifting 2-D pose estimation to 3D~\cite{zhao2017simple,wang20193d}. The progress in shape estimation is especially fueled by the development of SMPL~\cite{loper2015smpl}, a statistical low-dimensional representation of human body shape that enables end-to-end human shape estimation. HMR~\cite{kanazawa2018end} is a pioneering CNN based method that predicts human poses and shapes within the SMPL model from single RGB images. This is followed by~\cite{pavlakos2018learning,xu2019denserac} that further incorporate rendered silhouettes and texture maps for improved performance. Temporal information has also been exploited by~\cite{humanMotionKanazawa19,kocabas2019vibe} in inferring full-body poses and shapes from videos.

\textbf{Event camera and applications.} 
As an emerging bio-inspired imaging sensor, event camera~\cite{gallego2019event} is different from conventional frame cameras in many ways. 
The most important concept of event camera is an event, represented as a triplet, $(\mathbf{x}, t, p)$, which measures a noticeable change of brightness of at a specific pixel location $\mathbf{x}$, the time of occurrence $t$, and its polarity status $p$, which only occurs when either increasing or decreasing brightness exceeds a preset threshold.
Rather than capturing images at a fixed frame rate, events are asynchronously registered at per-pixel level in event camera. 
The stream of events is also spatially much sparser comparing to conventional frame cameras, where each image is densely packed with a full stack of per-pixel values.
Hence event camera is capable of perceiving local motions in the scene as a stream of sparse and asynchronous events.
Owing to its unique advantages of high temporal resolution, low latency, high dynamic range, and low power consumption, event camera has found applications in a growing list of computer vision tasks, including camera pose estimation~\cite{gallego2017event}, feature tracking~\cite{gehrig2018asynchronous}, optical flow~\cite{zhu18evflownet}, multi-view stereo~\cite{rebecq2018emvs}, gesture recognition~\cite{amir2017low}, motion deblurring~\cite{jiang2020learning}, among others. 
%

Meanwhile, there has been little investigation into event camera based estimation of 3D articulation of human body pose and shape. The efforts of DHP19~\cite{calabrese2019dhp19} and EventCap~\cite{xu2020eventcap} are perhaps the most related. %
In DHP19~\cite{calabrese2019dhp19}, a CNN model is devised to estimate 2-D human pose; it unfortunately cannot output 3D pose and shape.
EventCap~\cite{xu2020eventcap} aims to capture 3D human motions, which however demands more than just event signals: a stream of gray-scale images is also a necessary part of the input.
Its workflow starts with a pre-trained CNN-based pose and shape detection module that takes as input a stream of low-frequency gray-scale images; the detected results are then used as initial estimates to infill the intermediate poses to reconstruct high-frequency motion details constrained by event trajectories, following~\cite{gehrig2018asynchronous}; The poses are further refined using the silhouette information gathered from the events.
Compared with EventCap, our work includes the optical flow inferred from events to alleviate the requirements of a stream of gray-scale images over time, which enables our method to estimate 3D human pose and shape from events as the major input data source feasible.

\section{Our Approach}

\textbf{Event camera} provides a stream of event signals, with an event being a triplet of $(\mathbf{x}, t, p)$. At the same time, event camera typically outputs low-frame-rate gray-scale images. In our work, the stream of events is decomposed into a sequence of $N$ event packets, $\mathcal{E}=\{\mathcal{E}_{t_i}\}_{i=1}^N$, where an event packet $\mathcal{E}_{t_i}$ represents the set of events collected from time $t_{i-1}$ to $t_{i}$, as shown in Fig.~\ref{fig:pipeline}. Then, we divide each events packet $\mathcal{E}_{t_i}$ into $M$ sequential subsets in temporal order, and each subset is aggregated to be one channel of the event frame $I_{t_i}$. \cite{gehrig2019end} Thus an event frame $I_{t_i}$ consists of $M$ channels which are in temporal order. Intuitively, our representation of temporal channels is simple yet effective to include temporal information for human pose estimation. Besides, the event camera is assumed to be static and its intrinsic parameters are known.

\textbf{Human shape} is represented by the SMPL model~\cite{loper2015smpl}. SMPL model is a differentiable function, denoted by $\mathbf{v}=\mathcal{M}(\boldsymbol{\beta}, \boldsymbol{\theta})\in\mathbb{R}^{6,890\times3}$. Given the shape parameters $\boldsymbol{\beta}$ and pose parameters $\boldsymbol{\theta}$, the model outputs a triangular mesh with 6,890 vertices. The shape parameters $\boldsymbol{\beta}\in\mathbb{R}^{10}$ are the linear coefficients of a PCA shape space that mainly determines individual body features such height, weight and body proportions. The PCA shape space is learned from a large dataset of body scans. $\boldsymbol{\theta}\in\mathbb{R}^{72}$ are the pose parameters that mainly describe the articulated pose, consisting of one global rotation of the body and the relative rotations of 24 joints in axis-angle representation. Finally, the human shape is produced by first applying shape-dependent and pose-dependent deformations to the template body, then using forward-kinematics to articulate the template body shape to its target pose, and deforming the surface mesh by linear blend skinning. At the same time, the 3D and 2-D joint positions, $\mathbf{J}_{3D}$ and $\mathbf{J}_{2D}$, can be obtained by linear regression from the output mesh vertices and projection of 3D joints. 

Our method, EventHPE, is summarized in Fig.~\ref{fig:pipeline}, which consists of two stages. 
(i) The first stage, described in Sec.~\ref{sec:optical-flow-prediction}, is the optical flow inference from events. The event stream is first converted into a sequence of event frames and the event frames are fed into a CNN model to predict the corresponding optical flows. 
(ii) The second stage, described in Sec.~\ref{sec:pose-estimation}, is the estimation of shapes through time. A sequence of event frames together with corresponding optical flows are passed through a CNN model to extract their vectorized feature representations, and then fed into an RNN model to estimate the pose variations and global translation variations across each time interval. In this work, we assume that the beginning pose and shape is known. If not, we can extract the beginning pose and shape by pre-trained CNN-based models, such VIBE~\cite{kocabas2019vibe}, similarly to the previous work~\cite{xu2020eventcap}. Note that we \emph{only} need the pose and shape at the beginning time or for the first frame of gray-scale image. Finally, the estimated shapes through time can be obtained accordingly.

\subsection{Unsupervised Learning of Optical Flow}
\label{sec:optical-flow-prediction}
Based on the observation where the two types of modalities, events and optical flow, are closely related to human motions on an image, we propose to introduce optical flow inferred from events in our method to provide more explicitly geometric clues. Using the event frame $I_{t_i}$ as the input, we train a CNN model, denoted by FlowNet, to predict the optical flow $O_{t_i}$. The FlowNet model is like an encoder-decoder architecture and can be trained by unsupervised learning. The loss functions used to train the model, similar to~\cite{zhu18evflownet}, include a photo-metric and a smoothness loss where photo-metric loss describes the pixel-intensity difference between warped and target image, and smoothness loss describes the difference between each in-pixel flow with its neighboring pixels' flows. More details can be found in supplementary materials.

\subsection{Pose and Shape Estimation}\label{sec:pose-estimation}
For each time interval $(t_{i-1}, t_{i})$, the event frame $I_{t_i}$ and its corresponding optical flow $O_{t_i}$ are concatenated together, and then fed into a CNN model to extract its vectorized feature representation. A sequence of these temporal features is passed through a GRU model to obtain the desired outputs, including inter-frame pose variations $\Delta \boldsymbol{\hat \theta}_{t_i}\in \mathbb{R}^{144}$ and global translation variations $\Delta \mathbf{\hat d}_{t_i}\in \mathbb{R}^{3}$ for each time interval. After articulating the beginning pose and shape, we can obtain the estimated shapes sequentially. 

Specifically, the predicted global translation $\mathbf{\hat d}_{t_i}$ at time $t_i$ can be obtained by $\mathbf{\hat d}_{t_i}=\mathbf{\hat d}_{t_{i-1}} + \Delta \mathbf{\hat d}_{t_i}$, which leads to the translation loss,
\begin{equation}
    \label{eq:translation-loss}
    \mathcal{L}_{\text{trans}} = \sum_{t_i} \|\mathbf{d}_{t_i} - \mathbf{\hat d}_{t_i}\|^2_2,
\end{equation}
where $\mathbf{d}_{t_i}$ is the target global translation at time $t_i$.

As for the predicted pose $\boldsymbol{\hat \theta}_{t_i}$ at time $t_i$, instead of using the 3D axis-angle representation of relative rotations of 24 joints as the pose in SMPL model, we propose to use 6D representation of rotations as the pose, which shows better performance for estimating human pose than axis angles~\cite{zhou2019continuity}. The $j$-th relative rotation at time $t_i$ is given by
\begin{equation}
    \label{eq:pose-time-point}
    \boldsymbol{\hat \theta}^{j}_{t_i} = \mathbf{R}^{-1}\big(\mathbf{R}(\Delta \boldsymbol{\hat \theta}^{j}_{t_i})\mathbf{R}(\boldsymbol{\hat \theta}^{j}_{t_{i-1}})\big),
\end{equation}
where $\mathbf{R}(\cdot)$ is the function that transforms the 6D rotational representation to the $3\times3$ rotation matrix. Instead of using Euclidean distance, we propose to use geodesic distance in $SO(3)$ to measure the distance between the predicted and target poses, which is defined as
\begin{equation}
    \mathcal{L}_{\text{pose}} = \sum_{t_i}\sum_{j} \arccos^2\Big(\frac{\Tr\big(\mathbf{R}(\boldsymbol{\theta}^{j}_{t_i})^{\top}\mathbf{R}(\boldsymbol{\hat \theta}^{j}_{t_i})\big)-1}{2}\Big).
\end{equation}
We also consider the position errors of 3D and 2-D joints, which are given by
\begin{align}
    \label{eq:joints-loss}
    \mathcal{L}_{\text{3D}} &= \sum_{t_i} \sum_{j} \|\mathbf{J}^j_{3D, t_i} - \mathbf{\hat J}^j_{3D, t_i}\|^2_2,\\
    \mathcal{L}_{\text{2D}} &= \sum_{t_i} \sum_{j} \|\mathbf{J}^j_{2D, t_i} - \pi(\mathbf{\hat J}^j_{3D, t_i})\|^2_2,
\end{align}
where $\mathbf{\hat J}^j_{3D, t_i}$ is the predicted 3D position of joint $j$ and $\pi(\cdot)$ is the projection function. 

\begin{figure}[]
    \centering
    \includegraphics[width=\columnwidth]{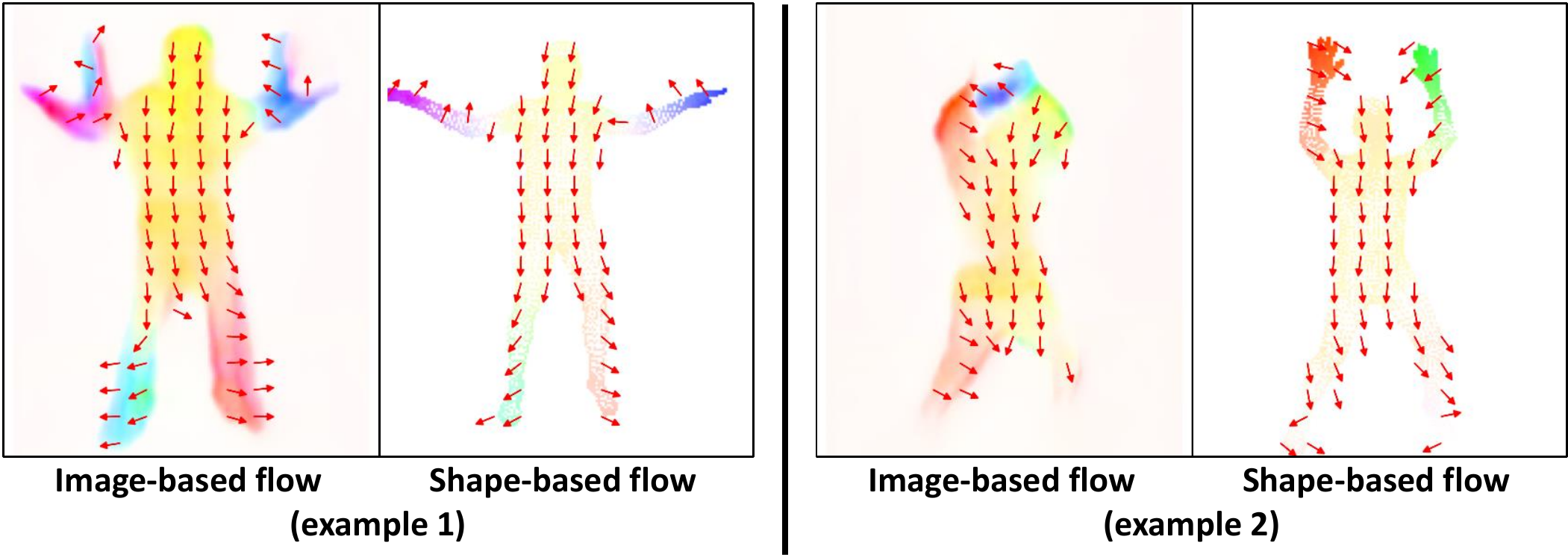}
    \caption{An illustration of image-based and shape-based flows. Image-based flow is the optical flow; shape-based flow refers to the movements of projected vertices of human body shape in image. The two flows are separately shown in color images, with red arrows displaying local flow directions. Note the pixel-level flow directions from the two modalities should be consistent.}
    \label{fig:flowloss-demo}
\end{figure}

Finally, we propose a novel coherence loss of two types of flows, \emph{image-based flow} (optical flow) and \emph{shape-based flow} (vertices movement of human body shape). Both flows originate from human motions, and the coherence loss ensures the consistency between the 2-D optical flow and 3D shape vertices flow, thus serving to regularize the motion estimation problem.
Specifically, the optical flow $O_{t_i}$ inferred from events in Sec~\ref{sec:optical-flow-prediction} is the image-based flow, as is shown in Fig.~\ref{fig:flowloss-demo}. The shape-based flow is obtained by projecting two sequential human body shapes on the image and calculating the movements of corresponding vertices, that is
\begin{equation}
    \mathbf{F}^{\text{shape}}_{t_{i}} = \pi(\mathbf{\hat v}_{t_{i}}) - \pi(\mathbf{\hat v}_{t_{i-1}}),
\end{equation}
where $\mathbf{F}^{\text{shape}}_{t_{i}}\in\mathbb{R}^{6890\times 2}$. Correspondingly, the image-based flow for the shape vertices can be obtained via bilinear sampling on the optical flow, which is defined as 
\begin{equation}
    \mathbf{F}^{\text{img}}_{t_{i}}=\text{BilinearSample}(O_{t_i}, \pi(\mathbf{\hat v}_{t_{i-1}})).
\end{equation}
Then the coherence loss is defined as the cosine-distance between two types of flows as
\begin{equation}
    \mathcal{L}_{\text{flow}} = \sum_{t_i}\sum_{v} \frac{\langle \mathbf{F}^{\text{shape}}_{t_{i}, v}, \mathbf{F}^{\text{img}}_{t_{i}, v}\rangle}{\|\mathbf{F}^{\text{shape}}_{t_{i}, v}\|_2 \cdot \|\mathbf{F}^{\text{img}}_{t_{i}, v}\|_2},
\end{equation}
where $v$ is the index of vertices and $\langle\cdot\rangle$ means the inner product.

In summary, we train the ShapeNet by minimizing the loss
\begin{align}
    \mathcal{L}=&\lambda_{\text{trans}}\mathcal{L}_{\text{trans}}+
    \lambda_{\text{pose}}\mathcal{L}_{\text{pose}}+\nonumber\\
    & \lambda_{\text{3D}}\mathcal{L}_{\text{3D}}+
    \lambda_{\text{2D}}\mathcal{L}_{\text{2D}}+
    \lambda_{\text{flow}}\mathcal{L}_{\text{flow}},
\end{align}
where $\lambda_{\text{pose}}, \lambda_{\text{trans}}, \lambda_{\text{3D}}, \lambda_{\text{2D}}$ and $\lambda_{\text{flow}}$ are the weights of corresponding losses.

\subsection{Our MMHPSD Dataset}

An in-house multi-modality dataset, MMHPSD, has been curated to facilitate empirical evaluation of our approach. It is also in response to the fact that the only existing dataset, EventCap~\cite{xu2020eventcap}, is not publicly available. 

\textbf{Data Acquisition.} In data acquisition stage, our multi-camera acquisition system contains 4 different imaging modalities, including one event camera, one polarization camera, and five RGB-D cameras. Specifically, the event camera is CeleX-V with resolution $1280\times 800$~\cite{chen2019celepixel}. \emph{Images from the frame-based cameras are all soft-synchronized with the gray-scale images of the event camera}; the events between two sequential gray-scale images are then collected synchronously. 15 subjects are recruited for data acquisition, with 11  being male and 4 being female. Each subject performs 3 groups of actions (21 different actions in total) for 4 times, where each group includes actions of fast/medium/slow speed respectively. Finally, 12 short video clips are collected from each subject, with each video having around 1,300 frames in 15 FPS. This amounts to 180 videos in total, with each video lasting about 1.5 minutes. The average number of events for the dataset is around 1 million per second. Overall, our dataset consists of 240k frames, where each frame contains a set of images: a gray-scale image, a sequence of inter-frame events, a polarization image, and five RGB and depth images. Further details regarding the dataset can be found in the supplementary.

\textbf{Annotation.} The SMPL shape and pose are annotated mainly based on the five RGB-D cameras, as follows. For each frame, the 2-D joints of all the RGB images are detected by OpenPose~\cite{cao2019openpose}; the depth of each 2-D joint is then obtained by warping its corresponding depth image to the RGB counterparts. After aggregating the five-view initial 3D pose by taking the average, we fit the SMPL male model to the initial pose via 3D SMPLify-x~\cite{pavlakos2019smplifyx} to obtain the initial SMPL parameters. For more precise annotations, the initial shape is fine-tuned by fitting to the point cloud collected from the five depth views using the L-BFGS algorithm~\cite{bollapragada2018LBFGS}, where the average distance of shape vertex to its nearest point in the point cloud is iteratively minimized~\cite{zuo2020sparsefusion, zuo2021unsupervised}. 
Exemplar annotated human shapes can be found in the supplementary.

\textbf{Dataset Comparison.} We compare our dataset with two existing event-based human motion datasets in Tab.~\ref{tab:dataset-compare}. Our dataset has the largest amount of frames and events. Although the number of subjects and sequences is not as many as DHP19, our dataset is multi-modality and accurately provides both SMPL pose and shape annotations, which possesses great potential in other research scenarios.

\begin{table}[]
    \centering
    \resizebox{\columnwidth}{!}{
    \begin{tabular}{c|ccccc}
    \bottomrule \hline
        Dataset & Seq\#/Sub\# & Frame\# & Pose & Shape & MM\\
    \hline
        DHP19 \cite{calabrese2019dhp19} & 33/17 & 87k & Yes & No & No \\
        EventCap \cite{xu2020eventcap} & 2/6 & - & No & No & No \\
        MMHPSD (ours) & 12/15 & 240k & Yes & Yes & Yes\\
    \hline \toprule 
    \end{tabular}}
    \caption{A tally of existing event-based human motion datasets, compared in terms of number of action sequences (Seq\#) per subject and number of subjects (Sub\#), number of frames (Frame\#), availability of annotated poses (Pose) and shapes (Shape), and multi-modality (MM).}
    \label{tab:dataset-compare}
\end{table}

\section{Experiments}

\begin{table*}[htb]
    \centering
    \resizebox{\textwidth}{!}{
    \begin{tabular}{c|p{105pt}|p{25pt}|p{60pt}p{60pt}p{60pt}p{55pt}p{40pt}}
    \bottomrule \hline
        \makecell[c]{} & \makecell[c]{Models} & \makecell[c]{Input} & \makecell[c]{MPJPE $\downarrow$} & \makecell[c]{PA-MPJPE $\downarrow$} & \makecell[c]{PEL-MPJPE $\downarrow$} & \makecell[c]{PCKh@0.5 $\uparrow$} & \makecell[c]{PVE $\downarrow$} \\
    \hline
        \cellcolor{white}  & \makecell[c]{DHP19~\cite{calabrese2019dhp19}} &  \makecell[c]{E} & \makecell[c]{80.08} & \makecell[c]{74.55} & \makecell[c]{131.73}  & \makecell[c]{0.80} & \makecell[c]{-}\\ 
        \cellcolor{white}  \multirow{-2}{*}{DHP19} & \makecell[c]{DHP19 + Flow} &  \makecell[c]{E+F} & \makecell[c]{76.76} & \makecell[c]{71.68} & \makecell[c]{130.37}  & \makecell[c]{0.82} & \makecell[c]{-}\\
    \hline
        \cellcolor{white} & \makecell[c]{HMR~\cite{kanazawa2018end}} & \makecell[c]{G} & \makecell[c]{-} & \makecell[c]{64.78} & \makecell[c]{95.32}  & \makecell[c]{0.61} & \makecell[c]{-}\\ 
        \cellcolor{white} & \makecell[c]{VIBE~\cite{kocabas2019vibe}} & \makecell[c]{V} & \makecell[c]{-} & \makecell[c]{50.86} & \makecell[c]{73.10}  & \makecell[c]{0.76} & \makecell[c]{-}\\
    \cline{2-8}
        \cellcolor{white} & \makecell[c]{EventCap(HMR)~\cite{xu2020eventcap}} & \makecell[c]{E+G} & \makecell[c]{-} & \makecell[c]{62.62} & \makecell[c]{89.95} & \makecell[c]{0.64} & \makecell[c]{-}\\
        \cellcolor{white} & \makecell[c]{EventCap(VIBE)~\cite{xu2020eventcap}} &  \makecell[c]{E+G} & \makecell[c]{-} & \makecell[c]{50.35} & \makecell[c]{71.85} & \makecell[c]{0.77} & \makecell[c]{-}\\
        \cellcolor{white} & \makecell[c]{DHP19~\cite{calabrese2019dhp19}} & \makecell[c]{E} &  \makecell[c]{72.42} & \makecell[c]{65.87} & \makecell[c]{74.04} & \makecell[c]{0.81} & \makecell[c]{-}\\
    \cline {2-8}
        \cellcolor{white} & \makecell[c]{EventHPE(HMR)} & \makecell[c]{E+F} & \makecell[c]{-} & \makecell[c]{53.72} & \makecell[c]{77.80}  & \makecell[c]{0.71} & \makecell[c]{-}\\ 
        \cellcolor{white} & \makecell[c]{EventHPE(VIBE)} & \makecell[c]{E+F} & \makecell[c]{-} & \makecell[c]{48.87} & \makecell[c]{69.58}  & \makecell[c]{0.79} & \makecell[c]{-}\\ 
        \cellcolor{white} \multirow{-8}{*}{MMHPSD} & \makecell[c]{EventHPE} &  \makecell[c]{E+F} & \makecell[c]{\textbf{71.79}} & \makecell[c]{\textbf{43.90}} & \makecell[c]{\textbf{54.96}} & \makecell[c]{\textbf{0.85}} & \makecell[c]{\textbf{53.90}}\\
    \hline
        \cellcolor{white} & \makecell[c]{EventHPE(w/o flow)} &  \makecell[c]{E} & \makecell[c]{80.99} & \makecell[c]{49.43} & \makecell[c]{60.90}  & \makecell[c]{0.82} & \makecell[c]{59.77}\\ 
        \cellcolor{white} & \makecell[c]{EventHPE(w/o flow loss)} &  \makecell[c]{E+F} & \makecell[c]{78.48} & \makecell[c]{47.36} & \makecell[c]{57.09}  & \makecell[c]{0.83} & \makecell[c]{56.58}\\ 
         \cellcolor{white} & \makecell[c]{EventHPE(w/o geodesic)} &  \makecell[c]{E+F} & \makecell[c]{77.29} & \makecell[c]{49.02} & \makecell[c]{60.55}  & \makecell[c]{0.83} & \makecell[c]{59.84}\\
        \cellcolor{white} \multirow{-3}{*}{Ablation} & \makecell[c]{EventHPE(w/o joints)} &  \makecell[c]{E+F} & \makecell[c]{73.79} & \makecell[c]{44.59} & \makecell[c]{55.91}  & \makecell[c]{0.84} & \makecell[c]{54.73}\\
    \hline \toprule  
    \end{tabular}
    }
    \caption{Quantitative evaluations on DHP19 and MMHPSD datasets, and the ablation studies on MMHPSD dataset. The input sources are represented by events (E), optical flow (F), gray-scale image (G) and video (V). The unit of joint errors is millimeter and PCK is proportional value.}
    \label{tab:pose-estimation}
\end{table*}

\begin{figure*}[]
    \centering
    \includegraphics[width=0.99\textwidth]{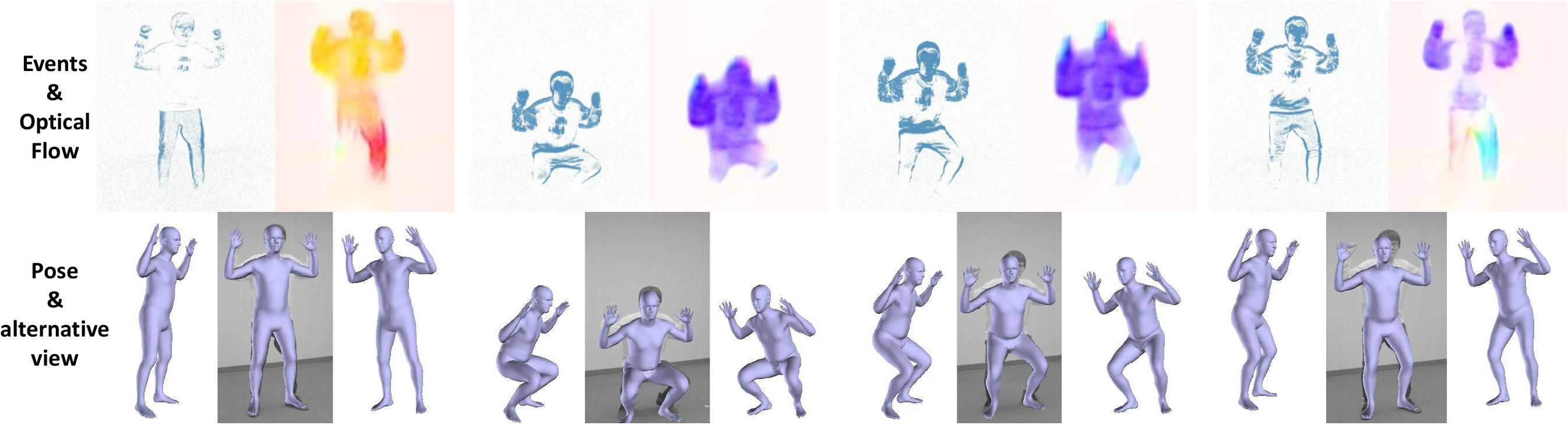}
    \caption{A sampled sequence of event frames and corresponding optical flows are shown in the first row; the second row shows each of the estimated shapes and two alternative views.}
    \label{fig:example}
\end{figure*}

In this section, we first describe the implementation details for training, and explain the evaluation metrics reported. Then, we compare our method with several event-based, frame-based and video-based approaches. Finally, we present the ablation studies to show the effectiveness of individual components in our method.

\textbf{Implementation Details.} During training, we choose inter-frame events as one event packet and then transform the packet into a 4-channel event frame ($M=4$) with each channel aggregating events for about 15 milliseconds. We also tried 1, 2, 8 and empirically find $4$ gives better results. Event frames are resized to $256\times256$. Note that when testing, we do not have such constraint, that is the temporal length of an event packet can be dynamic, depending on the speed of generated events, since fast motion often generates events much more faster than slow motion. ResNet50 \cite{he2016deep} is used as the backbone model of CNN, and one-layer GRU with 2048 hidden dimension is used as RNN in ShapeNet. The weights $\lambda_{\text{pose}}, \lambda_{\text{trans}}, \lambda_{\text{3D}}, \lambda_{\text{2D}}, \lambda_{\text{flow}}$ are set to $20, 10, 1, 10, 0.1$ respectively. The sequence length used in ShapeNet training and testing is $16$, which is a sequence of approximately $1$ second length. The batch size used to train the models is set to be $16$. The learning rate for FlowNet is $0.0001$ for $15$ epochs, and for ShapeNet is $0.001$ for $5$ epochs and decays to $0.0001$ for $3$ epochs. Models are trained on a single RTX 2080Ti.

\textbf{Evaluation.} Similar to previous works \cite{kocabas2019vibe,kanazawa2018end}, we report five different metrics, mean per joint position error (MPJPE), Procrustes-aligned MPJPE (PA-MPJPE), pelvis-aligned MPJPE (PEL-MPJPE), percent of correct key-points (PCKh@0.5) and per vertex error (PVE). PA-MPJPE compares predicted and target joints after rotation and translation alignment, while PEL-MPJPE compares after only translation alignment of two pelvis joints. For PCKh@0.5, the joint that had distance error less than 50\% of the head bone length after pelvis alignment is considered as correct key-point. The distance error of each vertex of SMPL mesh is used to calculate PVE.

\subsection{Empirical Results}
\label{sec:comparison-to-baselines}

\begin{figure*}[]
    \centering
    \includegraphics[width=0.88\textwidth]{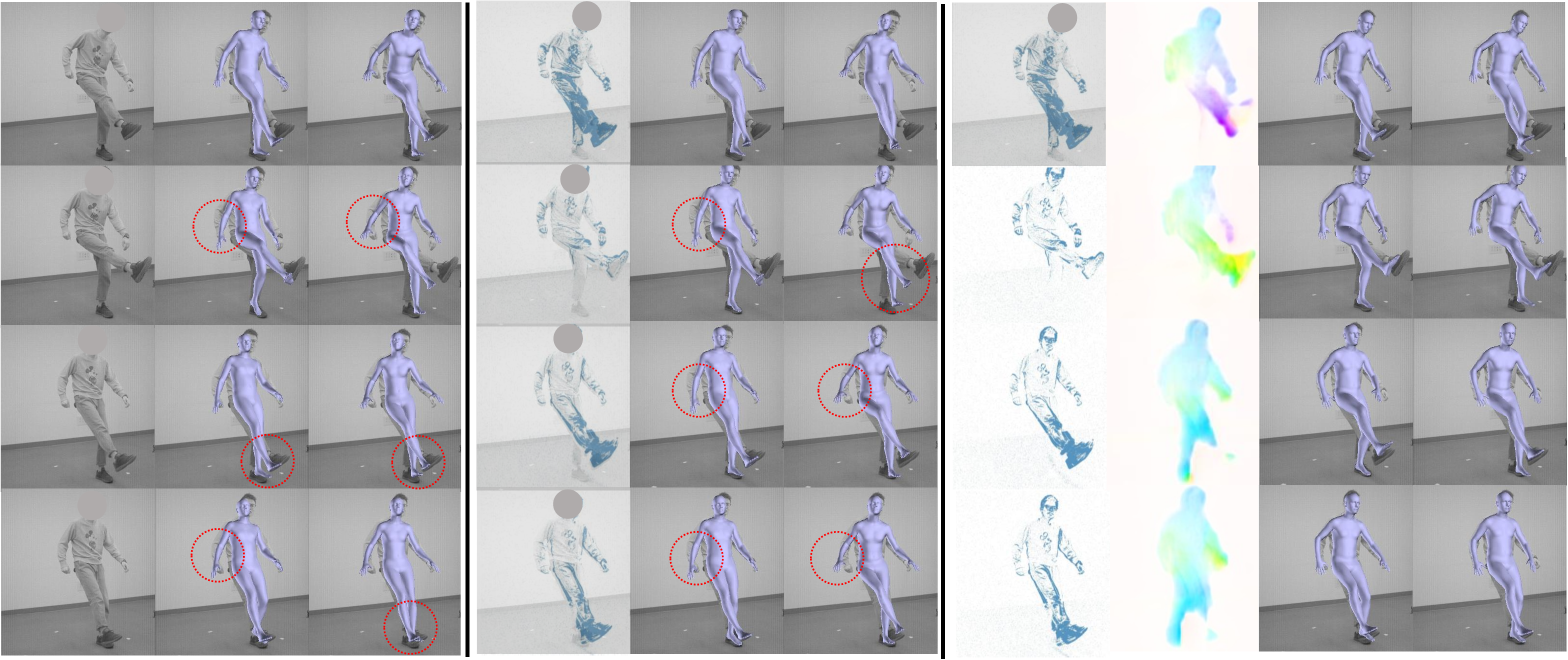}
    \includegraphics[width=0.88\textwidth]{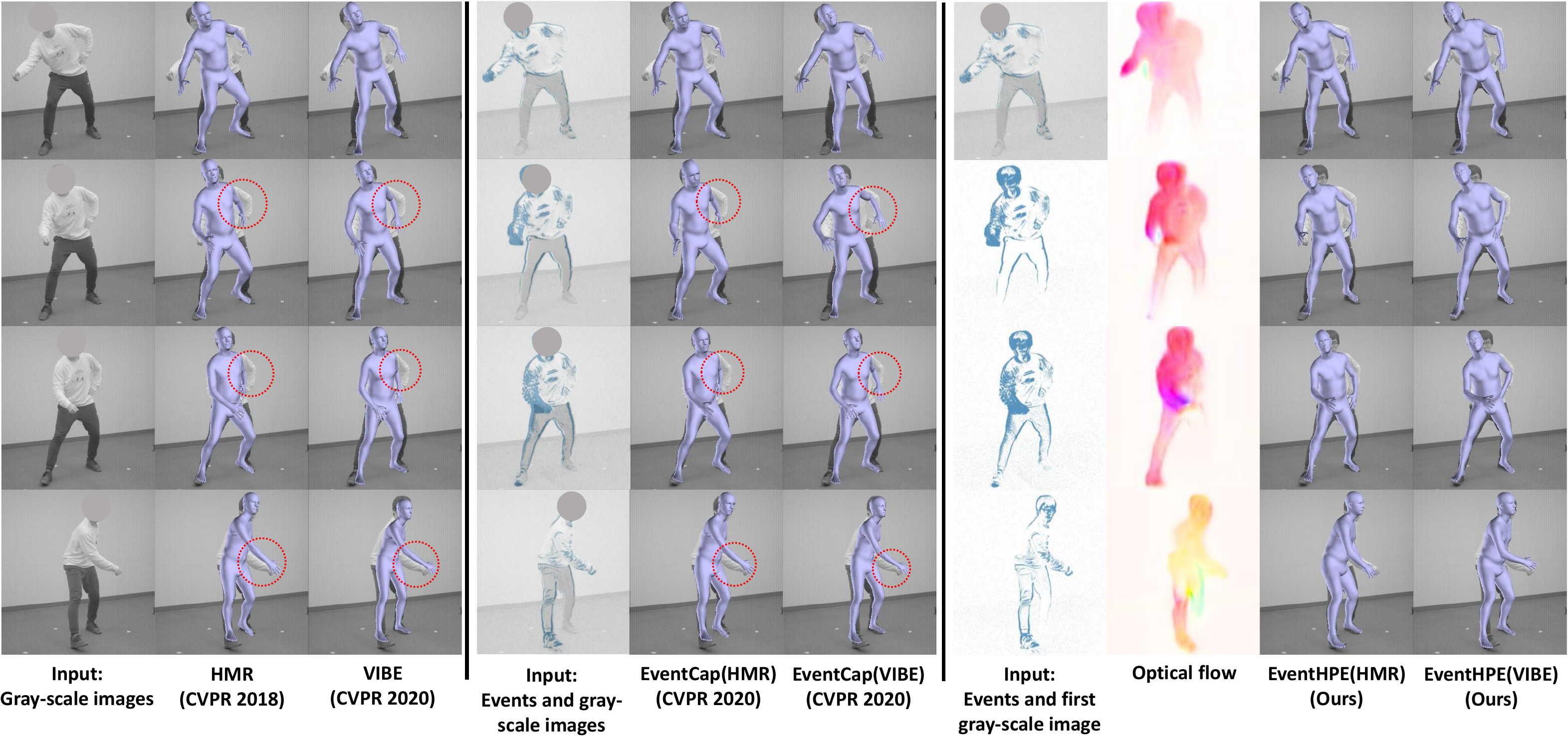}
    \caption{Qualitative results of the comparison to baselines. Our method is able to rectify the poses and shapes estimation through time \emph{even if the beginning pose and shape given by HMR or VIBE is not correct}. Note that our method only requires the pose and shape detection on the first frame of gray-scale image while EventCap requires on a sequence of images.}
    \label{fig:qualitative-results}
\end{figure*}

\textbf{DHP19 dataset} \cite{calabrese2019dhp19} only provides multi-view events stream and motion capture data of joints without gray-scale images and SMPL shape annotations. So we use this dataset to demonstrate the effect of optical flows in event-based pose estimation. We denote the method presented in \cite{calabrese2019dhp19} as \emph{DHP19}, and \emph{DHP19+Flow} means the input consists of both event frames and optical flows that are predicted by our FlowNet trained on MMHPSD dataset. The quantitative results in the first two rows of Tab.~\ref{tab:pose-estimation} show that the joint position error in terms of MPJPE and PA-MPJPE metrics decreases more than 3mm, while PEL-MPJPE does not show such a large increment. The reason is that we only detect 2-D joints instead of a whole body shape, which means that the pelvis translation alignment may cause larger distance error of other joints. The PCKh also shows consistent increase after optical flows are used as the other source of input data. The quantitative results on DHP19 dataset demonstrate the effect of our proposed idea to include optical flow to extract more explicit geometric information from events to help events-based pose estimation.

\textbf{MMHPSD dataset} provides a variety of data sources and well-aligned pose and shape annotations, which enables us to compare our event-based method with various baselines. Quantitative results are reported in Tab.~\ref{tab:pose-estimation} and qualitative results are shown in Fig.~\ref{fig:qualitative-results}.

\emph{HMR} \cite{kanazawa2018end} is used as a frame-based baseline, and \emph{VIBE} \cite{kocabas2019vibe} is applied as a video-based baseline. Note that HMR or VIBE predicts a weak camera model without global translation, and both of them use the neutral SMPL model, which is different from the male model used in MMHPSD. Therefore, the quantitative evaluations of MPJPE and PVE will not be reported.

Another three categories of event-based baselines are described as follows. \emph{DHP19} \cite{calabrese2019dhp19} is 2-D pose estimation method from events, and we assume that the ground-truth depths of detected 2-D joints are known for comparison in 3D. \emph{ EventCap(HMR)} or \emph{EventCap(VIBE)} means that HMR or VIBE is used in EventCap \cite{xu2020eventcap} to detect pose and shape on a sequence of gray-scale images as initial values. Since the authors have not published their code and evaluation dataset yet, we re-implement EventCap using PyTorch L-BFGS optimizer~\cite{bollapragada2018LBFGS,pytorch} and PyTorch3D differential render \cite{ravi2020pytorch3d}. As for our method, \emph{EventHPE(HMR)} or \emph{EventHPE(VIBE)} means HMR or VIBE is used to detect the beginning pose and shape on the first frame of gray-scale image, corresponding to the two cases of EventCap for comparison. \emph{EventHPE} represents the case where the ground-truth pose and shape is used as the beginning.

Since DHP19 uses the ground-truth depths of detected 2-D joints to obtain 3D pose, we compare it to our method EventHPE for fair comparison. The quantitative results show that our method yields more than 20mm decrease of joint errors of PA-MPJPE and PEL-MPJPE, while less than 1mm joint errors of MPJPE. This could attributes to the factors that our method predicts a whole body shape with the topology as the constraint while \emph{DHP19} only detects the individual joints independently. Therefore, our method gives much lower joint errors after alignment. 

\begin{figure*}[]
    \centering
    \includegraphics[width=0.9\textwidth]{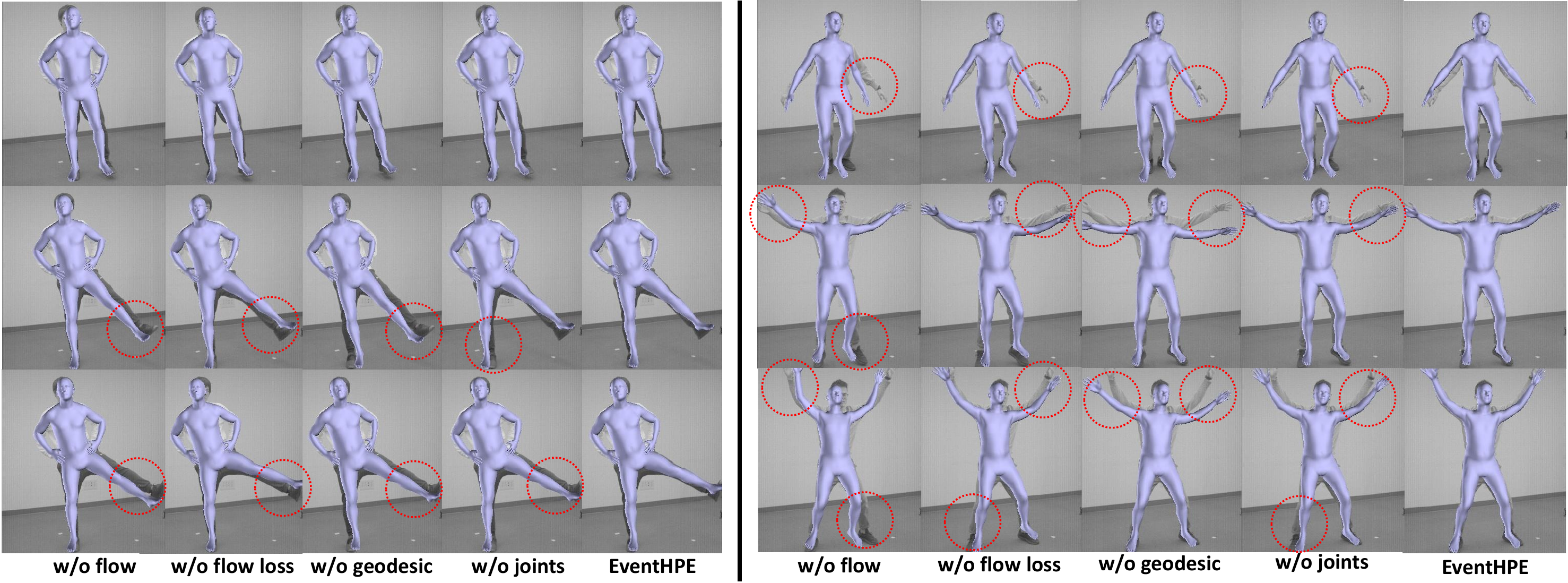}
    \caption{Qualitative results of ablation studies. Optical flow and geodesic distance are both important in inferring accurate shapes.}
    \label{fig:ablation}
\end{figure*}

For fair comparison of EventCap with our EventHPE, we can look at the quantitative results of EventCap(HMR) v.s. EventHPE(HMR) and EventCap(VIBE) v.s. EventHPE(VIBE). Note that our method only requires the extraction of pose and shape on the first frame of gray-scale image as priors while EventCap requires the extraction on all the gray-scale images over time. Though the accuracy of priors affects the performances of both methods, there are still noticeable improvements on joint errors or PCK. In the case of HMR, EventCap(HMR) presents about 2mm reduction of PA-MPJPE, 5mm reduction of PEL-MPJPE and 0.03 growth of PCK over HMR, while our method EventHPE(HMR) yields improvements of about 11mm, 17mm and 0.1 respectively, which are more than triple compared with EventCap(HMR). A similar trend can be observed in the case of VIBE. EventHPE(VIBE) presents about 2mm of PA-MPJPE, 3.5mm of PEL-MPJPE and 0.03 of PCK improvements over VIBE, and EventCap(VIBE) improves only 0.5mm , 1.5mm and 0.01 respectively. Though the improvements of performance in the case of VIBE is not as high as that of HMR, which may attribute to the factor that the performance of HMR allows a larger space of improvement than VIBE, the joint errors or PCK of EventCap and EventHPE in the case VIBE are consistently better than in the case of HMR. This observation show that the priors can affect the overall performance of both method, but our method is more robust than EventCap. The reason can be attributed to the setting where events are the only source of input given the beginning pose and shape, that is, events and the optical flows (predicted from events) provide a better chance to rectify the future prediction when incorrect beginning pose is given. On the contrary, EventCap is constrained to rectify by the incorrect initial estimates on the whole sequence of gray-scale images. 

The qualitative results in Fig.~\ref{fig:qualitative-results} also demonstrate the effectiveness of our method. Eight examples are sampled from two test sequences and displayed in Fig.~\ref{fig:qualitative-results}. We can see that even if the beginning pose and shape estimated by HMR or VIBE, displayed in the first row of each sequence, is not accurate or cannot align well with the subjects in the image, EventHPE can still rectify the following predictions to give well-aligned poses and shapes given the events and corresponding inferred optical flow. EventCap, however, can only slightly adjust the estimates as it requires the estimates of HMR or VIBE on every gray-scale image in a sequence, which constrains the pose adjustments to the in-correct ones across time. 

\subsection{Ablation Study}


In this section, we conduct ablation studies to evaluate individual components in our method. \emph{EventHPE(w/o flow)} means optical flow and flow coherence loss are not used in our method, \emph{EventHPE(w/o flow loss)} means optical flow is used as the input while flow coherence loss is not included, \emph{EventHPE(w/o geodesic)} means Euclidean distance of pose is use during training instead of geodesic distance, and \emph{EventHPE(w/o joint)} means joints supervision is not used in training. The quantitative and qualitative results are shown in Tab.~\ref{tab:pose-estimation} and Fig.~\ref{fig:ablation}. When comparing these four ablation models with EventHPE, a key observation is that the joint errors and shape vertex error will increase about 6mm for the model without optical flow or geodesic distance, 3-4mm for the model without flow coherence loss, while only 1-2mm for the model without joints loss during training. The qualitative results in Fig.~\ref{fig:ablation} further demonstrate that optical flow or geodesic distance plays an important role in our method. The model without optical flow or geodesic distance presents worse alignment of human body with the latent gray-scale images, which shows that geodesic distance is better than Euclidean distance to measure human pose distance in $SO(3)$ and optical flow with flow coherence loss can provide more explicit geometric information in human shape estimation from events.

\section{Conclusion and Outlook}
In this paper, we present an approach to estimate 3D human shapes from event camera. Empirical evaluations demonstrate the applicability and effectiveness of our method. A potential limitation of our work is that we need the beginning pose and shape provided or detected on the first frame of gray-scale images. 
For future work, we will focus on address the problem of inferring 3D shapes solely from event signals.

\section*{Acknowledgement}
Thank all the volunteers who contribute to the dataset, and thank Shuang Wu and Wei Ji for their constructive advice. This work is supported by the University of Alberta Start-up grant, the the UAHJIC grants, and the NSERC Discovery Grant No. RGPIN-2019-04575.

\newpage
{\small
\bibliographystyle{ieee_fullname}
\bibliography{egbib}
}

\newpage
\appendix










\section{Unsupervised Learning of Optical Flow}
The FlowNet can be trained by unsupervised learning via warping loss of two sequential gray-scale images $(I_{t_{i-1}}, I_{t_i})$. The loss functions used to train the model, similar to \cite{zhu18evflownet}, includes a photo-metric and a smoothness loss.

Given the predicted optical flow $O_{t_i}$, the warped image $\hat I_{t_i}$ can be obtained by warping the second image $I_{t_i}$ to the first image $I_{t_{i-1}}$ via bilinear sampling. The photo-metric loss describes the difference between $I_{t_{i-1}}$ and $\hat I_{t_i}$,
\begin{align}
    \label{eq:photo-metric-loss}
    & \mathcal{L}_{\text{photo}}(u, v; I_{t_{i-1}}, I_{t_i}, O_{t_i}) = \nonumber\\
    & \sum_{x, y} \rho (I_{t_{i-1}}(x, y) -  I_{t_i}(x+u(x, y), y+v(x, y))),
\end{align}
where $\rho$ is the Charbonnier loss function defined as $\rho(x)=\sqrt{x^2+\epsilon^2}$ and $(u, v)$ is the 2D direction of the predicted flow $O_{t_i}$. The Charbonnier loss is more robust than the absolute difference. The smoothness loss constraints the output flow by minimizing the difference of each in-pixel flow and its neighboring flows,
\begin{align}
    \label{eq:smoothness-loss}
    &\mathcal{L}_{\text{smooth}}(u, v; O_{t_i}) =  \nonumber\\
    & \sum_{x, y} \sum_{i,j\in \mathcal{N}(x, y)} \rho(u(x, y) -  u(i, j)) + \rho(v(x, y) -  v(i, j)),
\end{align}
where $\mathcal{N}(x, y)$ is the neighbors of pixel $(x, y)$.

To summarize, FlowNet is trained by minimizing the loss
\begin{align}
    \label{eq:flow-training-loss}
    \mathcal{L}_{\text{optical-flow}} = \mathcal{L}_{\text{smooth}} + \mathcal{L}_{\text{photo}}.
\end{align}

\section{MMHPSD Dataset Details}
\textbf{Data Acquisition.} In data acquisition stage, our multi-camera acquisition system has 12 cameras out of 4 different types of imaging modality: one event camera, one polarization camera, five-view RGB-D cameras. \emph{All the frame-based cameras are soft-synchronized with the gray-scale images of the event camera} and the events between two sequential gray-scale images are collected synchronously. 15 subjects are recruited for the data acquisition, where 11 are male and 4 are female. Each subject is required to perform 3 groups of actions (21 different actions in total, as is shown in Fig.~\ref{tab:dataset-action}) for 4 times, where each group includes actions of fast/medium/slow speed respectively. 

\begin{figure*}[]
    \centering
    \includegraphics[width=0.74\textwidth]{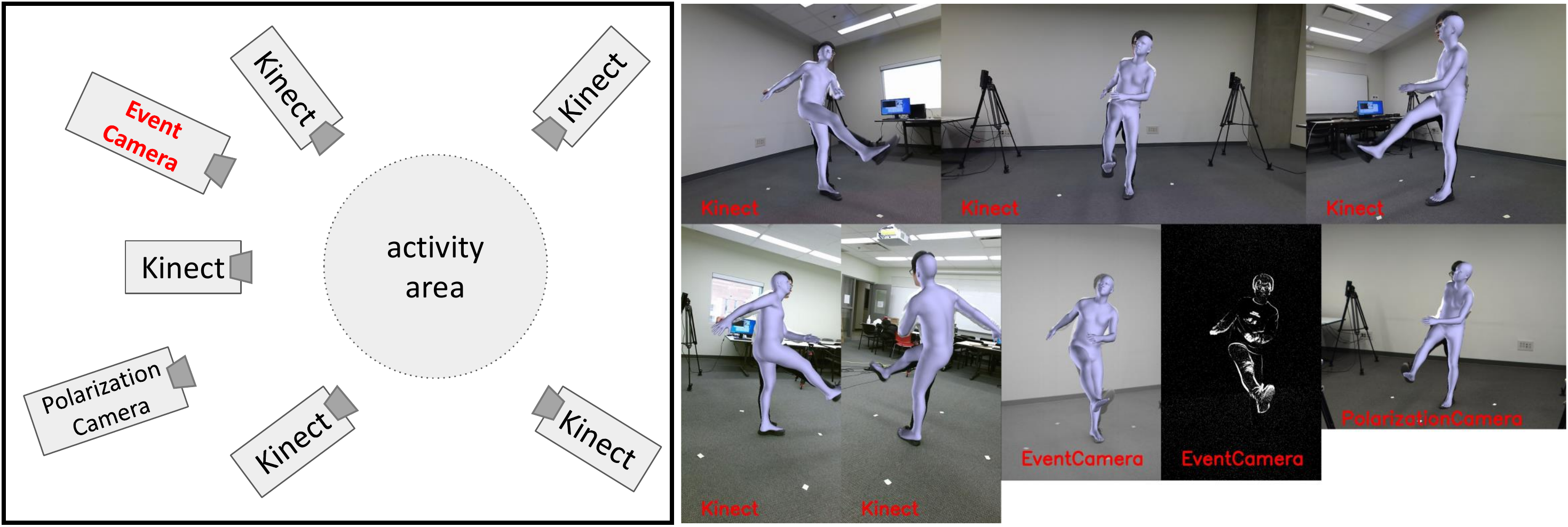}
    \includegraphics[width=0.8\textwidth]{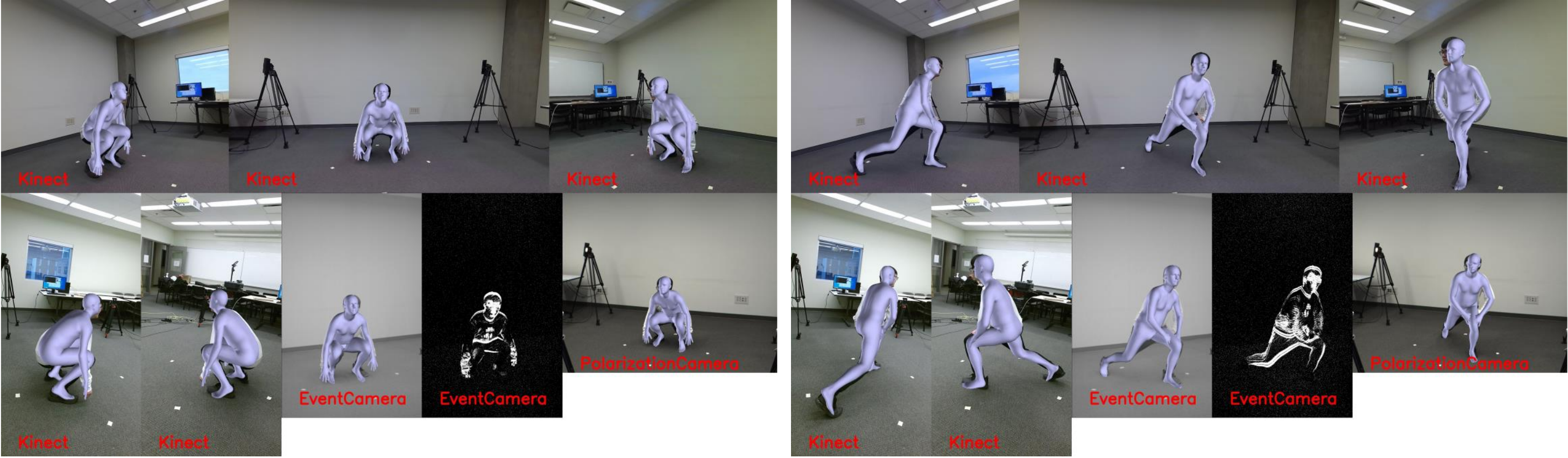}
    \caption{Layout of multi-camera acquisition system and three examples of annotated shapes rendered on multi-view images. The top left figure is the layout, and the other three figure present three examples of pose and shape annotation.}
    \label{fig:dataset-overview}
\end{figure*}

Finally, we collect 12 short videos for each subject and each video has around 1,300 frames with 15 FPS, that is 180 videos in total with each video lasting about 1.5 minutes. We conduct the annotations for each video and check manually whether the annotated shape aligns well with multi-view images. We abandon the unsatisfactory annotated shapes. Details on the number of frames per subject and number of annotated frames per subject are presented in Tab.~\ref{tab:dataset-details}. The event camera used is CeleX-V~\cite{chen2019celepixel} with resolution 1280x800 and the sensor frequency is 20-70 MHz. The MIPI interface supports up to 2.4Gbps transfer rate while the parallel interface supports the maximum readout of 140M pixels/second. The average number of events for the dataset is around 1 million per second. Fig.~\ref{fig:dataset-overview} presents the layout of our multi-camera system and three annotated shapes as examples. Overall, our dataset consists of 240k frames with each frame including a gray-scale image and inter-frame events, a polarization image, five-view color and depth images. 

\begin{table*}[]
    \centering
    \begin{tabular}{|c|c|c|}
    \bottomrule \hline
        group & speed & actions \\
    \hline
        1 & medium & jumping, jogging, waving hands, kicking legs, walk\\
        2 & fast & boxing, javelin, fast running, shooting basketball, kicking football, playing tennis, playing badminton\\
        3 & slow & warming up elbow/wrist ankle/pectoral, lifting down-bell, squating down, drinking water\\
    \hline \toprule  
    \end{tabular}
    \caption{Types of actions in each group. Subjects are required to do each group of actions for 4 times. The order of actions each time is random.}
    \label{tab:dataset-action}
\end{table*}

\begin{table*}[]
    \centering
    \begin{tabular}{|c|c|c|c|c|}
    \bottomrule \hline
        \makecell[c]{subject} & \ gender\  & \makecell[c]{\ \ \# of original\ \ \\\  frames\ } & \makecell[c]{\ \ \# of annotated\ \ \\\  frames\ } & \makecell[c]{\ \ \# of discarded\ \ \\\  frames\ } \\
    \hline
        1 & male & 15911 & 15911 & 0 (0.0\%) \\
        2 & male & 15803 & 15803 & 0 (0.0\%) \\
        3 & male & 16071 & 16071 & 0 (0.0\%)\\
        4 & male & 16168 & 16152 & 16 (0.01\%)\\
        5 & male & 16278 & 16262 & 16 (0.01\%)\\
        6 & male & 16715 & 16384 & 331 (2.0\%)\\
        7 & female & 16091 & 16091 & 0 (0.0\%)\\
        8 & male & 16257 & 15642 & 715 (4.4\%)\\
        9 & male & 15467 & 15461 & 6 (0.03\%)\\
        10 & male & 16655 & 16655 & 0 (0.0\%) \\
        11 & male & 16464 & 16443 & 21 (0.13\%)\\
        12 & male & 16186 & 16186 & 0 (0.0\%)\\
        13 & female & 16064 & 14562 & 1502 (9.4\%)\\
        14 & female & 15726 & 15166 & 560 (3.6\%)\\
        15 & female & 14193 & 14075 & 118 (0.8\%)\\
    \hline
        total & - & 240049 & 236764 & 3285 (1.4\%)\\
    \hline \toprule
    \end{tabular}
    \caption{Detail number of frames for each subject and the number of frames that have annotated SMPL pose and shape.}
    \label{tab:dataset-details}
\end{table*}


\end{document}